\documentclass[10pt,twocolumn,letterpaper]{article}

\usepackage{multirow}
\usepackage[pagenumbers]{cvpr} 

\newcommand{\B}[1]{\color{blue}}

\usepackage{tcolorbox}
\definecolor{cvprblue}{rgb}{0.21,0.49,0.74}
\usepackage[pagebackref,breaklinks,colorlinks,allcolors=cvprblue]{hyperref}

\def\paperID{12484} 
\def\confName{CVPR}
\def\confYear{2026}

\newcommand{\datasetName}{\textit{CataComp}}
\newcommand{\methodName}{\textit{CataractCompDetect}}
\newcommand{\anonymousHospital}{\textit{Sankara Eye Hospital}}
\newcommand{\anonymousLocation}{\textit{Bengaluru, India}}

\title{\methodName: Intraoperative Complication Detection in Cataract Surgery}

\author{
Bhuvan Sachdeva$^{1, 2}$, 
Sneha Kumari$^{1, 2}$, 
Rudransh Agarwal$^{1}$\thanks{Work done during internship at Sankara},
Shalaka Kumaraswamy$^{1}$, 
\and
Niharika Prasad$^{1}$,
Raphael Lechtenboehmer$^{3}$,
Simon Mueller$^{3}$,
Maximilian W. M. Wintergerst$^{3}$, 
\and
Thomas Schultz$^{3}$, 
Kaushik Murali$^{1}$\thanks{Corresponding authors: kaushik@sankaraeye.com and mohja@microsoft.com.},
Mohit Jain$^{2}$\footnotemark[2]
\\
\small $^{1}$Sankara Eye Hospital, Bengaluru, India\\
\small $^{2}$Microsoft Research, Bengaluru, India\\
\small $^{3}$University of Bonn, Bonn, Germany
}

\begin{document}
\maketitle
\begin{abstract}
Cataract surgery is one of the most commonly performed surgeries worldwide, yet intraoperative complications such as iris prolapse, posterior capsule rupture (PCR), and vitreous loss remain major causes of adverse outcomes. 
Automated detection of such events could enable early warning systems and objective training feedback. 
In this work, we propose \methodName{}, a complication detection framework that combines phase-aware localization, SAM 2-based tracking, complication-specific risk scoring, and vision-language reasoning for final classification.
To validate \methodName{}, we curate \datasetName{}, the first cataract surgery video dataset annotated for intraoperative complications, comprising 53 surgeries, including 23 with clinical complications.
On \datasetName{}, \methodName{} achieves an average F1 score of $70.63\%$, with per-complication performance of $81.8\%$ (Iris Prolapse), $60.87\%$ (PCR), and $69.23\%$ (Vitreous Loss).
These results highlight the value of combining structured surgical priors with vision-language reasoning for recognizing rare but high-impact intraoperative events. 
\end{abstract}
    
\section{Introduction}
\label{sec:intro}

Cataract surgery is one of the most common surgical procedures worldwide, with over 26 million surgeries performed annually~\cite{cicinelli2023cataracts}. While modern surgical techniques have achieved a high degree of standardization, intraoperative complications may still occur due to surgeon-specific variability and patient-related factors. Such complications pose significant risks to expected vision outcomes and, in severe cases, may result in permanent vision loss. Consequently, accurate and timely detection of such events is critical for improving surgical outcomes and enhancing patient care. In addition, automated complication detection can facilitate objective surgeon skill assessment and provide targeted feedback to trainees and novice surgeons.

Despite its clinical importance, automated complication detection in cataract surgery remains largely unexplored, due to two key barriers: (a) the lack of publicly available datasets; and (b) the visually ambiguous and transient nature of complications. Unlike well-studied tasks such as surgical phase prediction and tool segmentation~\cite{grammatikopoulou2021cadis, pissas2021effective, zisimopoulos2018deepphase, sachdeva2024phase, mueller2025phase}, complication detection presents unique challenges---complications are rare, heterogeneous across surgeries, and occur in short temporal windows. Moreover, they are difficult to identify from individual frames, requiring temporal context for robust identification.

\begin{figure}[t]
    \centering
    \includegraphics[width=\linewidth]{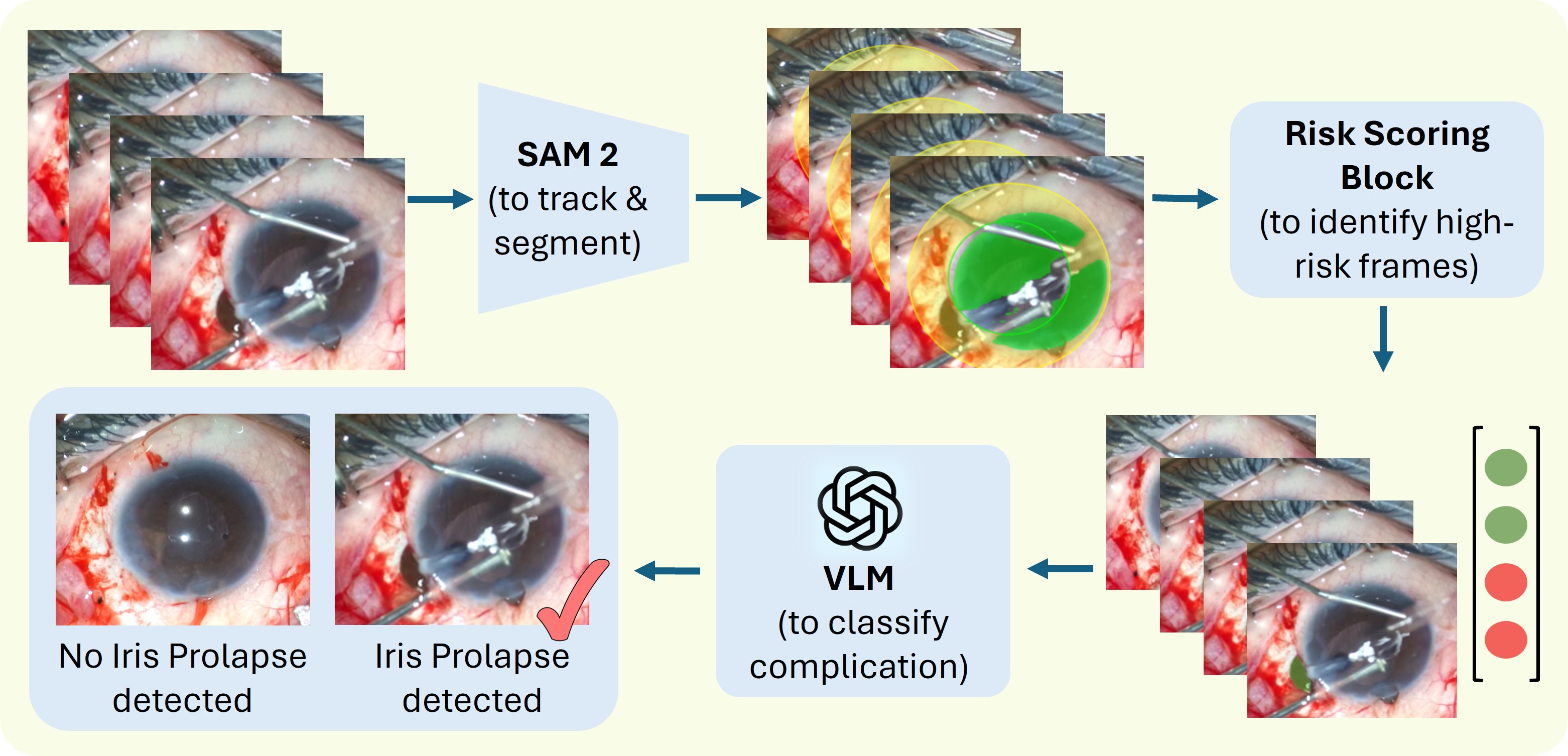}
    \caption{Overview of \methodName{} for complication detection in cataract surgery. It utilizes SAM 2 for tracking and segmentation, a Risk Scoring block to identify high-risk frames, and a Vision-Language Model (VLM) for final classification of complications like Iris Prolapse.}
    \label{fig:teaser}
\end{figure}

To address these challenges, we propose \methodName{} (Figure~\ref{fig:teaser}), a framework for automated detection of intraoperative complications in cataract surgery.
\methodName{} combines surgical context and anatomical cues with open-ended visual reasoning to achieve accurate localization and interpretable classification of three major complications. The framework first identifies relevant surgical phases using expert-derived priors (e.g., posterior capsular rupture typically occurs during cortical wash). Within these phases, a sliding-window based risk assessment module identifies visual anatomical patterns indicative of potential complications. Finally, a Vision-Language Model (VLM) classifies each high-risk segment as either containing the complication or not. The predictions across all segments are then aggregated to determine the final complication label for the video.

To evaluate \methodName{}, we curate \datasetName{}, a real-world clinical dataset at \anonymousHospital{}. \datasetName{} is the first dataset of cataract surgery videos annotated for intraoperative complications. It includes 53 surgeries (26m 3s$\pm$ 19m 14s), comprising of 23 complication cases across three clinically significant types: \textit{iris prolapse} (11), \textit{posterior capsule rupture (PCR)} (12), and \textit{vitreous loss} (12).
On the \datasetName{} benchmark, \methodName{} achieves an average F1-score of $70.63\%$ across these complications.
At the individual complication level, it detects iris prolapse, PCR, and vitreous loss,  with F1 scores of $81.8\%$, $60.87\%$, and $69.23\%$, respectively. 

\noindent \textbf{Key Contributions.} Our main contributions are as follows: 
\begin{itemize}
    \item We propose \textbf{\methodName{}}, a novel framework for intraoperative complication detection in cataract surgery that combines SAM~2-based anatomical tracking, complication-specific risk scoring, and vision-language reasoning to enable accurate and computationally efficient detection.
    \item We introduce \textbf{\datasetName{}}, the first video dataset specifically annotated for cataract surgery complications, establishing a benchmark for future research in this domain.
    \item We demonstrate that \methodName{} achieves strong performance across three clinically significant complications (\textit{Iris Prolapse}, \textit{PCR}, and \textit{Vitreous Loss}) with an average F1 score of 70.63\%.

\end{itemize}
\section{Related Work}

\subsection{Deep Learning in Cataract Surgery}
Deep learning methods have been applied to several tasks in cataract surgery, most prominently surgical phase recognition~\cite{morita2019real, zisimopoulos2018deepphase, fang2022global, mueller2025phase}, tool segmentation~\cite{grammatikopoulou2021cadis, pissas2021effective, zhao2022trasetr, sachdeva2024phase}, and surgeon skill assessment~\cite{giap2025catskill, hira2022video, kim2019objective}.
The CATARACTS dataset\footnote{\url{https://cataracts.grand-challenge.org/}} is a widely used benchmark for phase recognition, motivating numerous subsequent works. \citet{morita2019real} trained a CNN-based model to predict surgical phases at the frame level, while Deepphase~\cite{zisimopoulos2018deepphase} employed an RNN architecture to encode temporal aspects in tool usage for improved phase recognition.
However, these approaches relied on frame-level features, limiting their ability to capture long-term temporal dependencies. 
To address this, recent methods~\cite{czempiel2020tecno, fang2022global, mueller2025phase} leverage video-based features followed by multi-stage temporal CNNs to explicitly model temporal context, thereby achieving superior performance.

Parallel advances have been made in surgical tool segmentation in cataract surgery videos. \citet{grammatikopoulou2021cadis} introduced the CaDIS dataset as a benchmark for cataract surgical tool segmentation. Building on this, \citet{pissas2021effective} mitigated class imbalance in CaDIS through adaptive oversampling, while TraSeTR\cite{zhao2022trasetr} incorporated temporal context via tool tracking. More recently, \citet{sachdeva2024phase} demonstrated that integrating surgical phase priors further improves segmentation accuracy significantly.

While these tasks establish the foundation for spatio-temporal analysis in cataract surgery, higher level tasks such as complication detection and surgeon skill assessment demand a more nuanced understanding of surgical dynamics. \citet{hira2022video} and \citet{kim2019objective} modeled surgeon skill within specific surgical steps by conditioning on tool-tip trajectories, with the model trained using expert-assigned scores. CatSkill~\cite{giap2025catskill} introduced a rule-based framework that evaluates surgical skill by analyzing eye position throughout the procedure, including neutrality and centration metrics.
Although these works underscore the efficacy of deep learning in cataract surgery analysis, the problem of automated complication detection remains largely unexplored.

\subsection{Complication Detection in Surgical Videos}
\citet{tabuchi2022real} and \citet{morita2020real} formulated complication detection in cataract surgery as a frame-level binary classification task, distinguishing normal from abnormal frames using private datasets.
\citet{wang2022intelligent} proposed a step-recognition approach that leveraged the sequential order of surgical steps to identify abnormal procedures, effectively differentiating surgeries performed by novice and expert surgeons.

Beyond cataract surgery, prior work have explored complication detection in other surgical domains. For instance, in laparoscopic surgery, \citet{samuel2021unsupervised} developed an autoencoder-based unsupervised method on the Cholec80 dataset~\cite{cholec80}. However, anomalies were limited to superficial events such as bleeding, smoke, or camera occlusion, which are different from the clinically critical complications targeted in our work. Similarly, \citet{daneshgar2021visual} detected abrupt surgical tool movements to predict unintentional, tool-induced bleeding.

In summary, complication detection in surgical videos remains at an early stage.
Specifically, in cataract surgery, prior work have been limited to binary classification of complication versus non-complication videos. 
Our work extends this field by identifying real-world clinical complications and their specific types, 
thereby moving toward a more fine-grained and clinically meaningful understanding of intraoperative risks.
\section{Background}

\subsection{Cataract Surgery}
\begin{figure}[t]
    \centering
    \includegraphics[width=0.7\linewidth]{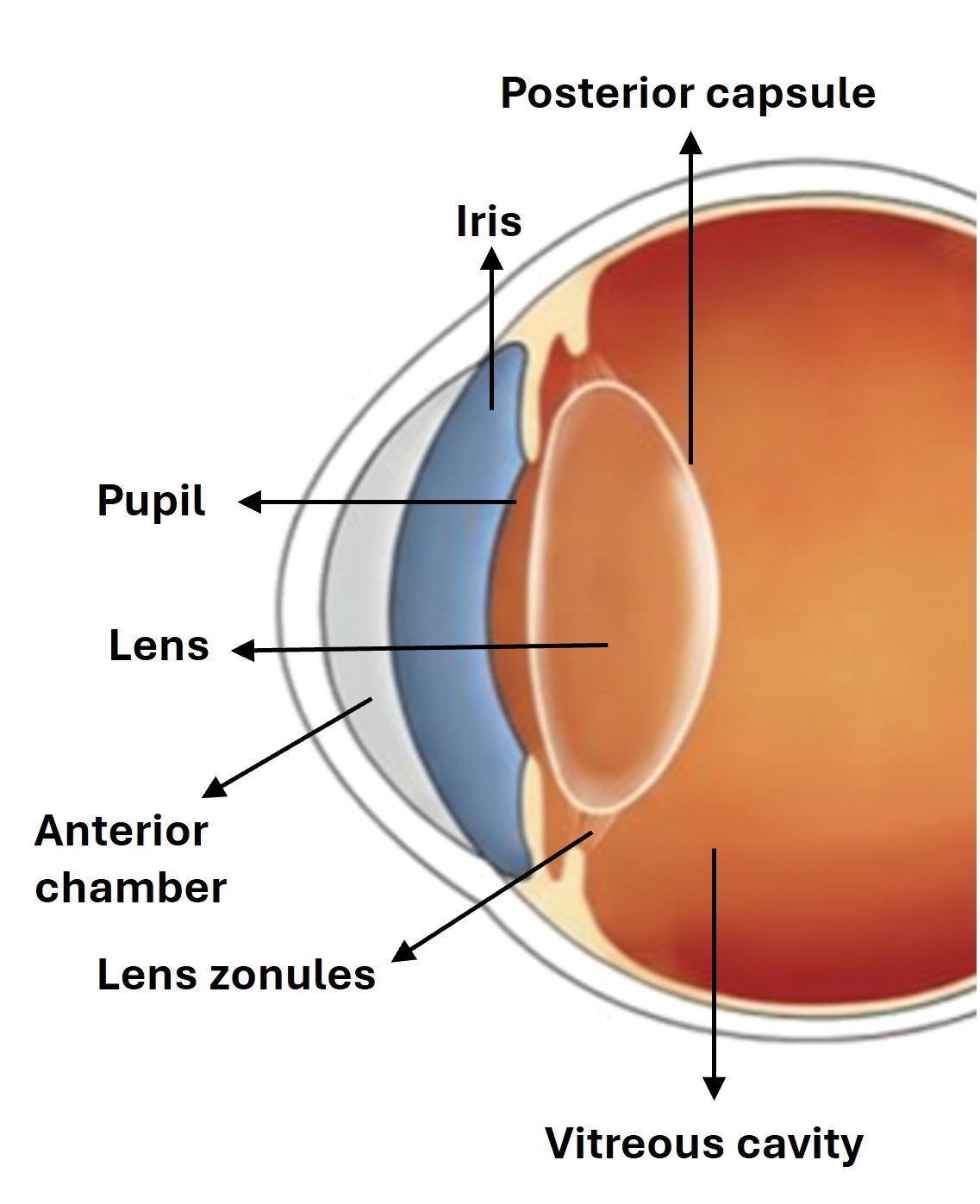}
    \caption{Anatomy of the eye}
    \label{fig:eye_anatomy}
\end{figure}

With advancing age, the natural lens of the eye (Figure~\ref{fig:eye_anatomy}) gradually loses transparency, leading to partial or complete vision loss. Cataract surgery restores vision by removing the opaque lens and replacing it with a clear artificial intraocular lens.

The two most widely used techniques for cataract surgery are Phacoemulsification and MSICS (Manual Small Incision Cataract Surgery). In phacoemulsification, a small corneal incision is created, and the dense central portion of the lens (the lens nucleus) is broken up and aspirated using an ultrasonic probe. This is followed by cortical wash to remove residual lens material, and finally, the intraocular lens is implanted. 
In resource-limited settings or in complex cases involving multiple pre-existing conditions, MSICS is more commonly performed. Unlike phacoemulsification, the lens nucleus is manually prolapsed through a slightly larger incision without the use of an ultrasonic probe. Consequently, MSICS serves as a low-cost alternative and is widely preferred in Low- and Middle-Income Countries (LMICs).
Although these procedures differ in incision size and nucleus management, they share several intraoperative steps and are susceptible to similar complications. In this work, we primarily focus on MSICS, as the work was conducted in an LMIC context.

\subsection{Complications in Cataract Surgery}
Despite standardized techniques and high success rates, cataract surgery remains vulnerable to intraoperative complications. These events can have serious consequences, including complete vision loss, underscoring the importance of accurate and timely detection.
In this work, we focus on three of the most frequent and clinically significant complications\footnote{Identified in consultation with four senior ophthalmologists from Asia and Europe}: \textit{iris prolapse}, \textit{posterior capsule rupture}, and \textit{vitreous loss}. These complications can occur in both phacoemulsification and MSICS due to their shared surgical steps. Figure \ref{fig:eye_anatomy} illustrates key ocular structures, including the iris, vitreous cavity, and posterior capsule, which are referenced below.

\begin{figure*}[t]
    \centering
    \includegraphics[width=0.9\linewidth]{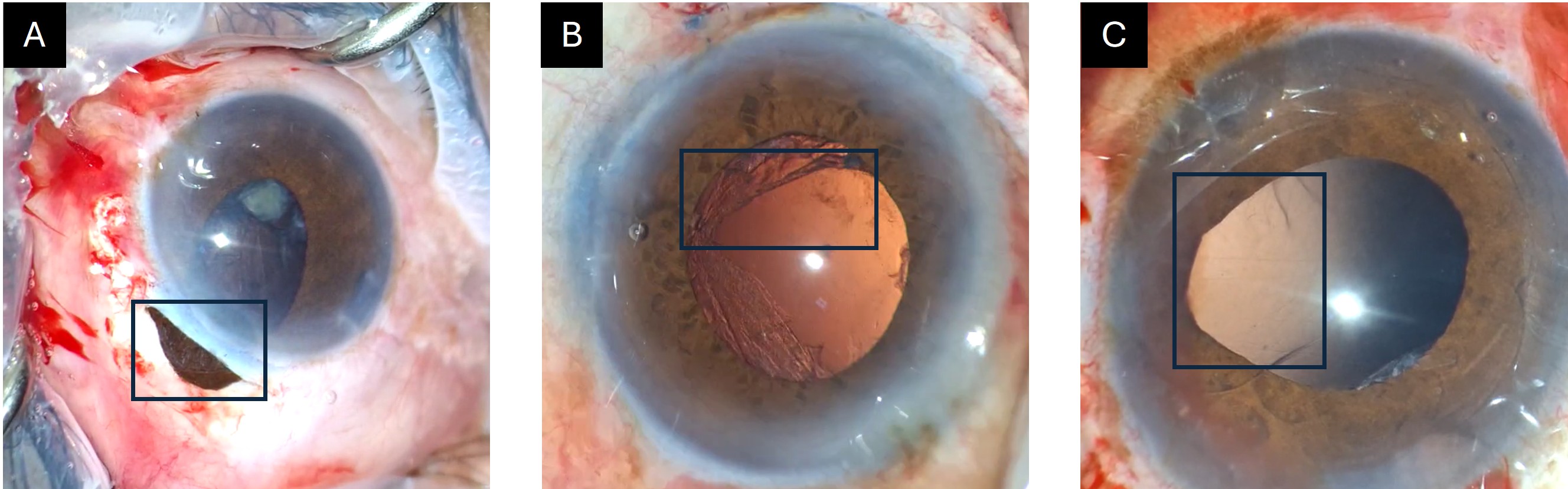}
    \caption{
    Examples of intraoperative anomalies in cataract surgery (highlighted by bounding boxes): 
    (A) \textbf{Iris Prolapse}: identified by brownish iris tissue protruding through the surgical incision, 
    (B) \textbf{Posterior Capsule Rupture (PCR)}: visible as a tear within the posterior capsule, and
    (C) \textbf{Vitreous Loss}: characterized by a forward bulge of the pupil and/or translucent web-like vitreous strands extending through the pupillary plane.
    }
    \label{fig:anomalies}
\end{figure*}

\textbf{Iris Prolapse} refers to iris tissue protruding through the surgical incision. It usually occurs due to subtle fluctuations in intraocular pressure or incision instability that allow the iris to slip outward. It appears as dark brownish-red iris tissue extending beyond its normal boundary (Figure~\ref{fig:anomalies}A). It requires immediate surgical intervention, typically iris repositioning, to prevent wound gaping, inflammation, or subsequent PCR. Iris prolapse may occur at various surgical phases, and delayed management can compromise incision integrity and visual outcomes.

\textbf{Posterior Capsule Rupture (PCR)} is a tear in the thin posterior capsule that supports the lens. It typically occurs during lens nucleus removal or cortical wash, when excessive pressure is exerted on the delicate capsule, causing it to rupture. It appears as a visible tear within the capsule (Figure~\ref{fig:anomalies}B). If unrecognized or inadequately managed, PCR can result in vitreous loss and intraocular lens displacement, increasing the risk of postoperative complications.

\textbf{Vitreous Loss} occurs when the gel-like vitreous prolapses from the vitreous cavity into the anterior chamber. It often follows a PCR or zonular dialysis (ZD)\footnote{Zonular Dialysis (ZD) is another known complication in cataract surgery; however, due to its relatively low incidence, it is outside the scope of this work.}, where a breach in the posterior capsule allows the vitreous gel to migrate outward. Visually, vitreous strands appear as translucent, web-like structures extending through the pupil or incision site, resulting in a forward bulging of the pupil (commonly referred to as a tear-shaped pupil) (Figure~\ref{fig:anomalies}C). Prompt recognition and management with anterior vitrectomy are essential to prevent further complications.

Note: PCR and vitreous loss are closely related.
However, not every PCR results in vitreous loss, whereas every vitreous loss typically occurs secondary to either PCR or ZD.
In \datasetName{}, all PCR cases led to vitreous loss, and no instances of ZD were observed, which is not fully representative of real-world surgical scenarios.

\section{Dataset}

\begin{table}[!ht]
\centering
\caption{\datasetName{} dataset statistics.}
\label{tab:dataset}
\begin{tabular}{l|cc}
\hline
\textbf{Complication} & \textbf{\# Videos} & \textbf{Average Duration} \\
\hline
None                & 32 & 18min 09s $\pm$ 07min 51s \\
Iris Prolapse       & 11 & 25min 38s $\pm$ 21min 47s \\
PCR                 & 12 & 50min 16s $\pm$ 23min 34s \\
Vitreous Loss       & 12 & 50min 16s $\pm$ 23min 34s \\
\hline
\textbf{Total}      & 53 & 26min 03s $\pm$ 19min 14s \\
\hline
\end{tabular}
\end{table}

\datasetName{} dataset comprises 53 MSICS videos with a total duration of 23hrs 0min 39sec (26m 3s $\pm$ 19m 14s per case), recorded at \anonymousHospital{} in \anonymousLocation{}. 
All videos were captured under routine surgical conditions using a smartphone-based recording setup. The videos were recorded at a resolution of $1920\times1080$ pixels and a frame rate of 30 frames per second.
Each video was annotated for three intraoperative complications: Iris Prolapse, PCR, and Vitreous Loss.
Annotations were performed at the video level by two resident ophthalmologists working collaboratively; labels were finalized through discussion and mutual consensus. Among all videos, 21 contain one or more complications. Detailed dataset statistics are provided in Table~\ref{tab:dataset}.

\section{Method}

\begin{figure*}[!t]
    \centering
    \includegraphics[width=\textwidth]{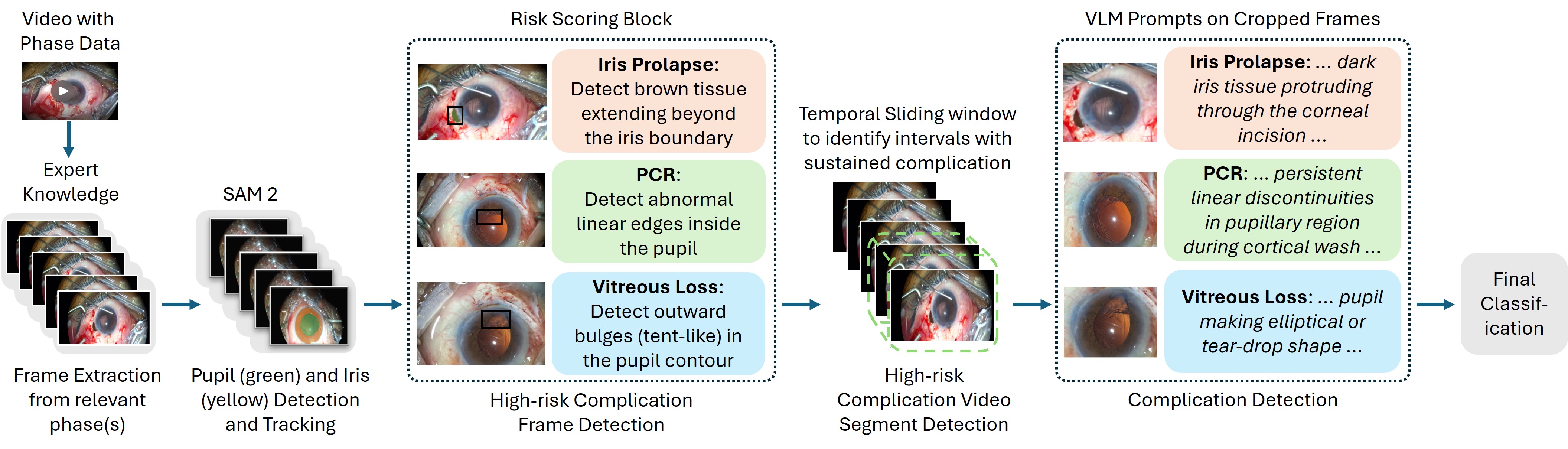}
    \caption{Overview of the \methodName{} pipeline. The framework integrates surgical phase-aware localization, SAM 2-based tracking of pupil and iris, complication-specific risk assessment, and vision-language reasoning to detect and classify intraoperative complications.}
    \label{fig:pipeline}
\end{figure*}

Our proposed video processing pipeline, \methodName{} (Figure~\ref{fig:pipeline}), comprises of three key stages: SAM 2-based tracking, complication-specific risk scoring, and VLM-based classification, which we describe below:

\subsection{Preliminary: Localizing surgical complications}
In this stage, we leverage ophthalmologists' domain knowledge to identify the surgical phase(s) in which each complication is most likely to occur.
PCR typically occurs during the \textit{lens nucleus removal} or subsequent \textit{cortical wash} phase. Since the pupil region is largely obscured by the lens during lens removal, detection is focused on the \textit{cortical wash} phase, where posterior capsule tears are most visible. 
Vitreous Loss generally results from either PCR or ZD, and is therefore most likely to occur during or after the \textit{cortical wash} phase. For this complication, we analyze the video segment spanning from the start of \textit{cortical wash} to the beginning of the \textit{artificial lens insertion}.
In contrast, Iris Prolapse can occur at any point during the surgery, and hence detection is performed across the entire video sequence.

\subsection{SAM~2-based Tracking}
In this stage, we employ the pretrained SAM~2 model~\cite{ravi2024sam} to track the iris and pupil masks throughout the surgical video. Point-based prompts are used to initialize the tracking, and additional prompts are added in video segments where the model's predictions appear inaccurate.


\subsection{Complication-specific Risk Scoring}
We compute segment-level risk scores for each complication as described below.

\textbf{Iris Prolapse}:
Iris Prolapse is characterized by dark brownish-red tissue protruding beyond the normal iris boundary. To detect it, the periphery of the iris is monitored for uncharacteristic regions resembling iris tissue. SAM~2's automatic mask generator is then applied to identify all candidate masks within this periphery. Post-processing removes false positives, such as instruments, specular reflections, and blood artifacts, using heuristics based on mask size and color. The area of the final candidate mask serves as the risk score, with frames lacking a valid mask assigned a score of zero. A temporal sliding window is then applied to identify intervals exhibiting sustained iris protrusion, which are flagged as high-risk segments.

\textbf{PCR}:
PCR typically appears as a persistent linear discontinuity within the posterior capsule in the pupillary region. To detect this, histogram equalization is first applied, followed by edge detection to enhance and isolate faint linear structures inside the pupil. Each detected edge is scored based on the diagonal length of its minimum enclosing bounding box, emphasizing long, straight edges over curved or noisy contours. 
The frame-level risk score corresponds to the longest such edge normalised with respect to the pupil area.
A temporal sliding window aggregates these scores over time to identify intervals with consistently long edges---patterns uncharacteristic of normal surgery. These intervals are flagged as high-risk segments.

\textbf{Vitreous Loss}:
Vitreous Loss typically manifests as an outward bulge or wedge-shaped deformation of the pupil contour. To capture this, the pupil boundary is partitioned into angular sectors (e.g., 0–30°, 30–60°) and the average radius within each sector is computed. Sectors exhibiting a consistently larger radius relative to adjacent regions indicate local outward protrusion. Each frame is assigned a risk score defined as the ratio between the maximum sector radius and the global mean radius. A temporal sliding window is then applied to identify intervals with sustained deviations, which are flagged as high-risk segments.

Each risk-scoring block follows a unified design: complications are defined by distinct visual cues, extracted through targeted image processing and evaluated temporally via sliding-window analysis. This design ensures reliable detection while remaining adaptable to the distinct visual characteristics of different complications. Its simplicity and modularity also make it easily extensible to other intra-operative events beyond those examined in this work.

\subsection{VLM-based Classification}
The final stage employed a Vision-Language Model (VLM) (GPT-5 in our case) to verify the presence of complications within the high-risk segments identified earlier. The top five high-risk segments from each video are selected where for each segment we send a representative frame for independently processed by the VLM. Each input consists of a cropped region encompassing the iris and pupil, accompanied by a complication-specific prompt that provides a concise yet detailed visual description of the target complication. This encourages the model to focus on clinically relevant features such as capsular tears, pupil deformation, and iris protrusion.

Prompts are crafted to guide the model's attention toward discriminative visual cues while grounding its reasoning in precise clinical definitions. For instance, key elements from the prompts include (see Supplementary for full prompt text):
\begin{itemize}[leftmargin=*, topsep=2pt, itemsep=2pt]
    \item \textit{Iris Prolapse}: ``... dark iris tissue protruding through the corneal incision and extending beyond the normal iris boundary...''
    \item \textit{PCR}: ``... PCR requires visible or strongly implied loss of the pupil contour -- a distinct discontinuity, irregular tear, or missing segment...''
    \item \textit{Vitreous Loss}: ``... Pupil peaking or elongation: Any deviation from circular toward ellipse or tear-drop; note apex sharpness and continuity of pupil margin...''    
\end{itemize}

This stage operates in a zero-shot setting, without any task-specific fine-tuning. Finally, predictions from high-risk segments are aggregated to assign the final complication label to each video.

\section{Experiments and Results}

\begin{table*}[ht]
    \caption{Performance of \methodName{} results on the \datasetName{} dataset.}
    \centering
    \begin{tabular}{l|cccc}
    \hline
    \textbf{Complication} & \textbf{Accuracy} & \textbf{Sensitivity} & \textbf{Specificity} & \textbf{F1 score} \\ \hline
    Iris Prolapse         & 92.45                            & 81.80                                    & 95.24                                    & 81.80                           \\
    PCR                   & 82.35                            & 58.33                                    & 89.74                                    & 60.87                           \\ 
    Vitreous Loss         & 84.31                            & 75.00                                    & 87.18                                    & 69.23                           \\ \hline
    \textbf{Average}      & 86.37                            & 71.71                                    & 90.72                                    & 70.63                           \\ \hline
    \end{tabular}
    \label{tab:main_results}
\end{table*}

\begin{table}[ht]
\centering
\caption{Effect of each stage on the number of videos passed forward and the number of predicted positives. Values in parentheses indicate the number of videos that truly contain the complication.}
\label{tab:stagewise_counts}
\begin{tabular}{l|ccc}
\hline
\textbf{Complication} &
  \textbf{Total} &
  \textbf{\begin{tabular}[c]{@{}c@{}}After Risk\\ Scoring\end{tabular}} &
  \textbf{\begin{tabular}[c]{@{}c@{}}After VLM\\ Classifier\end{tabular}} \\ \hline
Iris Prolapse & 53 (11) & 53 (11) & 11 (9) \\
PCR           & 51 (12) & 11 (7)  & 11 (7) \\
Vitreous Loss & 51 (12) & 32 (12) & 14 (9) \\ \hline
\end{tabular}
\end{table}

\subsection{Experimental Setup}
\textbf{Data and Compute.} We evaluate our proposed method on the \datasetName{} dataset. To balance the preservation of fine-grained anatomical detail with computational efficiency, all videos are processed at 5 frames per second and downsampled to a resolution of $960 \times 540$. All experiments are conducted on NVIDIA V100 GPUs with 32~GB of VRAM.

\textbf{Surgical Phase Localization.}
Expert-provided phase annotations are used to isolate the relevant surgical phases (cortical wash, lens nucleus removal, and artificial lens insertion) for PCR and Vitreous Loss. 
Two videos that do not contain the cortical wash phase are excluded from evaluation for PCR and Vitreous Loss.

\textbf{SAM 2-based Tracking.} For SAM~2-based tracking, each object (iris and pupil) is initialized using a single keyframe annotation containing two positive prompt points and one negative point placed near the periphery. These prompts enable the model to generate accurate segmentation masks, which are then propagated throughout the video. We use the SAM~2.1~H-Large pretrained checkpoint for tracking.

\textbf{Risk Scoring.} For risk scoring, we utilize a temporal sliding window of size 10 to identify high-risk segments. Adjacent high-risk segments produced by the risk scoring module are merged to eliminate redundant detections and ensure that the downstream VLM receives temporally diverse, non-overlapping candidate regions.

\textbf{VLM Classification.} GPT-5 is used for Iris Prolapse classification, whereas GPT-5 Mini is used for PCR and Vitreous Loss classification, with reasoning enabled for both models. Despite its larger size, GPT-5 exhibited substantially lower sensitivity for PCR and Vitreous Loss (Table~\ref{tab:ablation_vlm}), while GPT-5 Mini produced more stable and reliable predictions for these tasks.

\subsection{Results and Analysis}
We present the performance of \methodName{} method on the \datasetName{} dataset. Overall, \methodName{} achieves an average F1 score of $70.63\%$ across the three complications. Table~\ref{tab:main_results} reports video-level classification performance for each complication individually. The method performs best on Iris Prolapse (F1: $81.80\%$), followed by Vitreous Loss ($69.23\%$) and PCR ($60.87\%$), reflecting the inherently subtler and more short-lived visual signatures associated with PCR and Vitreous Loss.

To better understand the effect of each pipeline component, Table~\ref{tab:stagewise_counts} summarizes the progression of videos through the risk scoring and VLM-classification stages.

\textbf{Impact of Risk Scoring.}
The risk scoring module effectively prunes Normal videos while maintaining high sensitivity. For Vitreous Loss detection, the module filters the dataset from 51 to 32 videos by identifying only those containing high-risk segments. Notably, all 19 excluded videos correspond to Normal cases, indicating high specificity in early-stage filtering.
A similar trend is observed for PCR. The risk scoring module identifies 40 videos as Normal, 35 of which are true Normal cases. These results demonstrate the efficiency of the risk scoring mechanism in reducing the search space for the downstream VLM classifier.

\textbf{Impact of VLM-based classification.}
The VLM performs final classification on the high-risk segments identified earlier. Because Iris Prolapse produces distinct visual cues, a video is labeled positive whenever at least one segment is classified as positive. Under this rule, the VLM identifies 11 videos as positive, including 9 true cases.
PCR presents subtler cues, and thus a more sensitive decision rule is adopted: a video is marked positive if the VLM predicts either \emph{positive} or \emph{unsure} for any high-risk segment, and marked negative only if all segments are explicitly predicted as \emph{negative}.
For Vitreous Loss, 14 of the 32 videos are flagged as positive, 9 of which are true positives. A video is assigned a positive label if the VLM predicts positive with high confidence for at least one segment, or positive with medium confidence for at least two segments. 

\begin{table}[t]
\centering
\caption{Comparison of VLM models for complication classification. GPT-5 exhibits substantially lower sensitivity than GPT-5 Mini for both PCR and Vitreous Loss.}
\label{tab:ablation_vlm}
\begin{tabular}{l|l|ccc}
\hline
\textbf{Comp.} & \textbf{VLM} & \textbf{Sensitivity} & \textbf{Specificity} & \textbf{F1} \\ \hline
\multirow{2}{*}{\begin{tabular}[c]{@{}l@{}}Iris\\ Prolapse\end{tabular}} & GPT-5        & 81.80                & 95.24               & 81.80       \\
                     & GPT-5 Mini & 90.91 & 57.14 & 51.28 \\ \hline
\multirow{2}{*}{PCR} & GPT-5      & 41.67 & 97.44 & 55.56 \\
                     & GPT-5 Mini & 58.33 & 89.74 & 60.87 \\ \hline
\multirow{2}{*}{\begin{tabular}[c]{@{}l@{}}Vitreous\\ Loss\end{tabular}} & GPT-5        & 25.00                & 100.00               & 40.00       \\
                     & GPT-5 Mini & 75.00 & 87.18 & 69.23 \\ \hline
\end{tabular}
\end{table}

\begin{figure*}[t]
    \centering
    \includegraphics[width=0.9\linewidth]{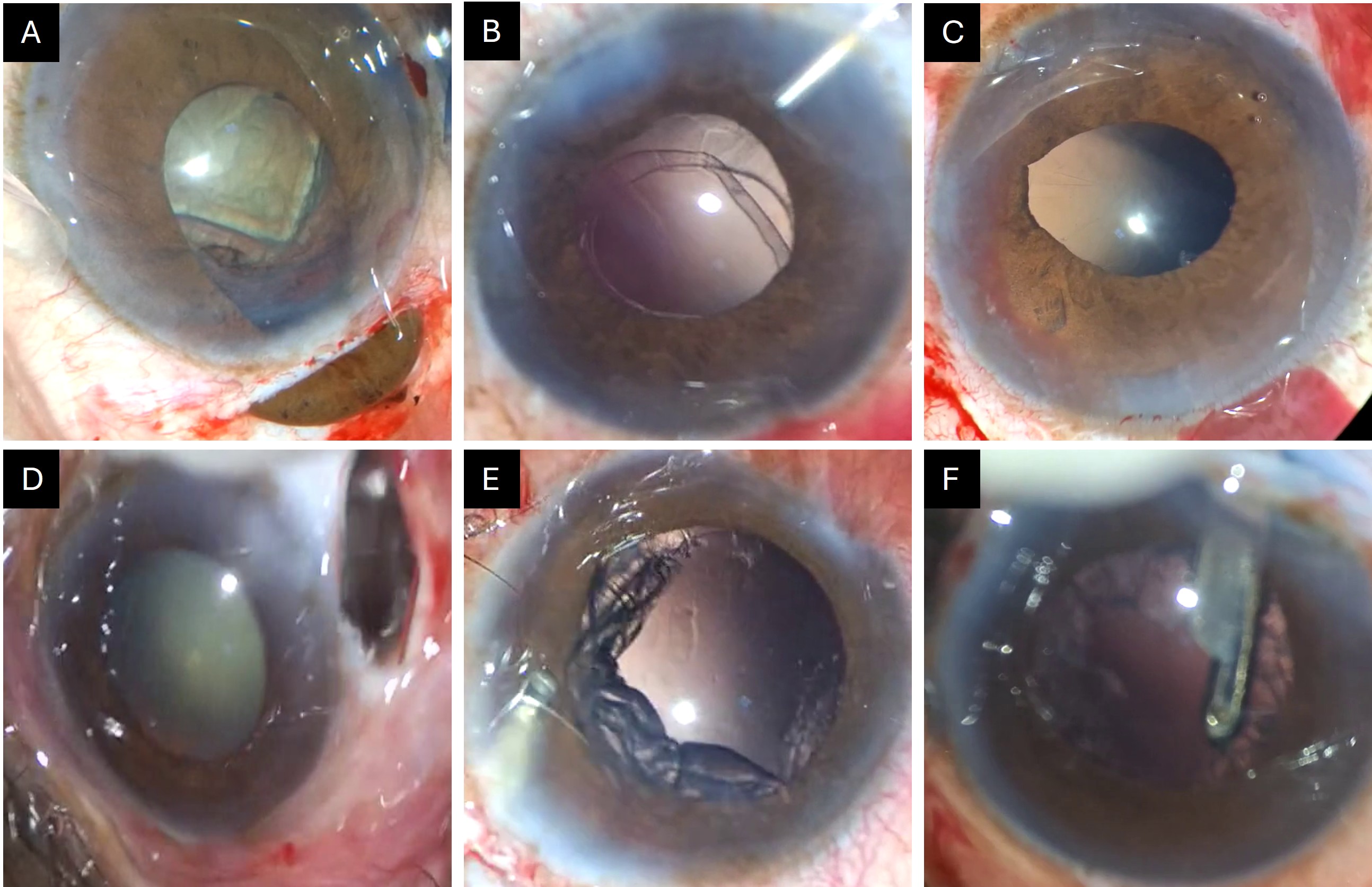}
    \caption{Representative frames that passed the Risk Scoring stage and were assigned a positive label by the VLM. True positives: Iris Prolapse (A), PCR (B), and Vitreous Loss (C). False positives: Iris Prolapse (D), PCR (E), and Vitreous Loss (F).}
    \label{fig:qual}
\end{figure*}

\textbf{Qualitative Analysis}.
Figure~\ref{fig:qual} demonstrates representative frames from video segments that were finally classified as containing a complication. The top row contains shows true positive cases, whereas the bottom row illustrates false positives. For instance, Figure~\ref{fig:qual}D shows the tip of a surgical tool that was misclassified as prolapsed iris tissue. Similarly, Figure~\ref{fig:qual}E shows residual cortex material remaining after lens extraction, which can visually resemble a torn posterior capsule. In Figure~\ref{fig:qual}F, the pupil boundary is poorly defined, which was mistakenly interpreted as pupil deformation. This resulted in a false positive classification for vitreous loss. The figure also highlights the complexity and similarities in the surgical scene, which makes detecting surgical complications challenging

\section{Discussion}

\textbf{Limitations.} Our study has several limitations.
First, the evaluation relies on a relatively small dataset. This is due to multiple factors: the nascent state of research on cataract surgery complication detection, the absence of publicly available datasets, the challenges inherent in recording and anonymizing surgical videos, and the resource-intensive process of obtaining complication annotations from ophthalmologists.
Second, our work lacks quantitative comparisons of \methodName{} with prior baseline methods. This is because no existing datasets or prior techniques specifically address automated detection of individual cataract surgery complication types, making direct benchmarking infeasible.
Third, our framework is optimized for cataract surgery and depends on complication-specific risk-scoring modules. While these tailored components contribute substantially to performance, extending the pipeline to detect complications in other surgical domains would require re-designing these specialized modules, thus limiting its generalizability.

\textbf{Future work.}
Future research should build upon the current framework to advance the field of intraoperative complication detection in cataract surgery. A key next step is the curation of a larger, more diverse dataset to evaluate the generalizability of emerging methods. Expanding the dataset with additional videos and annotatiions---across multiple complications and at different levels of granularity (such as video-level, phase-level, and frame-level)---would significantly broaden the scope of the field.
Such a dataset would also enable
exploration of learning-based methods, which were not feasible in our work due to limited data.
These methods may be less dependent on cataract-specific priors and could facilitate detecting complications across other surgical domains, supporting greater generalization.

Another compelling research direction is finetuning of VLMs.
Because surgical video data is inherently out-of-distribution for general-purpose VLMs, domain-adapted fine-tuning may improve robustness and reduce reliance on handcrafted components. This could also lead to the development of smaller, more efficient VLMs suitable for on-premise deployment, which is an essential requirement for clinical integration given the stringent data privacy and security constraints in healthcare settings.


\section{Conclusion}
In this work, we introduced \methodName{}, a novel framework designed for the challenging task of intraoperative complication detection in cataract surgery videos. Our proposed pipeline combines SAM~2-based tracking, complication-specific risk scoring, and vision-language reasoning to enable accurate and computationally efficient detection of clinically significant complications.
We also contribute \datasetName{}, the first real-world video dataset specifically annotated for cataract surgery complications, which we expect will serve as an important resource for advancing research in surgical complication detection. On \datasetName{}, \methodName{} achieves an average accuracy of $86.37\%$ and an average F1 score of $70.63\%$ across three complications. Performance varies by complexity and visual clarity of each complications, with F1 scores of $81.8\%$ for Iris Prolapse, $60.87\%$ for PCR, and $69.23\%$ for Vitreous Loss.
These results demonstrate that combining structured surgical priors, risk scoring, and vision-language reasoning offers a practical and effective strategy for complication detection, even in settings with limited annotated data.
We hope that this work lays the foundation for more comprehensive and real-time intraoperative monitoring systems capable of improving surgical safety and training.
{
    \small
    \bibliographystyle{ieeenat_fullname}
    \bibliography{main}

@article{grammatikopoulou2021cadis,
  title={CaDIS: Cataract dataset for surgical RGB-image segmentation},
  author={Grammatikopoulou, Maria and Flouty, Evangello and Kadkhodamohammadi, Abdolrahim and Quellec, Gwenol{\'e} and Chow, Andre and Nehme, Jean and Luengo, Imanol and Stoyanov, Danail},
  journal={Medical Image Analysis},
  volume={71},
  pages={102053},
  year={2021},
  publisher={Elsevier}
}

@article{mueller2025phase,
  title={Phase recognition in manual Small-Incision cataract surgery with MS-TCN++ on the novel SICS-105 dataset},
  author={Mueller, Simon and Sachdeva, Bhuvan and Prasad, Singri Niharika and Lechtenboehmer, Raphael and Holz, Frank G and Finger, Robert P and Murali, Kaushik and Jain, Mohit and Wintergerst, Maximilian WM and Schultz, Thomas},
  journal={Scientific Reports},
  volume={15},
  number={1},
  pages={1--10},
  year={2025},
  publisher={Nature Publishing Group}
}

@InProceedings{sachdeva2024phase,
    author="Sachdeva, Bhuvan
    and Akash, Naren
    and Ashraf, Tajamul
    and M{\"u}ller, Simon
    and Schultz, Thomas
    and Wintergerst, Maximilian W. M.
    and Singri, Niharika
    and Murali, Kaushik
    and Jain, Mohit",
    title="Phase-Informed Tool Segmentation for Manual Small-Incision Cataract Surgery",
    booktitle="Medical Image Computing and Computer Assisted Intervention -- MICCAI 2025",
    year="2026",
    publisher="Springer Nature Switzerland",
    address="Cham",
    pages="446--455",
    isbn="978-3-032-05114-1"
}

@article{tabuchi2022real,
  title={Real-time artificial intelligence evaluation of cataract surgery: a preliminary study on demonstration experiment},
  author={Tabuchi, Hitoshi and Morita, Shoji and Miki, Masayuki and Deguchi, Hodaka and Kamiura, Naotake},
  journal={Taiwan Journal of Ophthalmology},
  volume={12},
  number={2},
  pages={147--154},
  year={2022},
  publisher={Medknow}
}

@article{morita2020real,
  title={Real-time surgical problem detection and instrument tracking in cataract surgery},
  author={Morita, Shoji and Tabuchi, Hitoshi and Masumoto, Hiroki and Tanabe, Hirotaka and Kamiura, Naotake},
  journal={Journal of Clinical Medicine},
  volume={9},
  number={12},
  pages={3896},
  year={2020},
  publisher={MDPI}
}

@inproceedings{zisimopoulos2018deepphase,
  title={Deepphase: surgical phase recognition in cataracts videos},
  author={Zisimopoulos, Odysseas and Flouty, Evangello and Luengo, Imanol and Giataganas, Petros and Nehme, Jean and Chow, Andre and Stoyanov, Danail},
  booktitle={International conference on medical image computing and computer-assisted intervention},
  pages={265--272},
  year={2018},
  organization={Springer}
}

@article{morita2019real,
  title={Real-time extraction of important surgical phases in cataract surgery videos},
  author={Morita, Shoji and Tabuchi, Hitoshi and Masumoto, Hiroki and Yamauchi, Tomofusa and Kamiura, Naotake},
  journal={Scientific reports},
  volume={9},
  number={1},
  pages={16590},
  year={2019},
  publisher={Nature Publishing Group UK London}
}

@article{samuel2021unsupervised,
  title={Unsupervised anomaly detection for a smart autonomous robotic assistant surgeon (saras) using a deep residual autoencoder},
  author={Samuel, Dinesh Jackson and Cuzzolin, Fabio},
  journal={IEEE Robotics and Automation Letters},
  volume={6},
  number={4},
  pages={7256--7261},
  year={2021},
  publisher={IEEE}
}

@article{wang2022intelligent,
  title={Intelligent cataract surgery supervision and evaluation via deep learning},
  author={Wang, Ting and Xia, Jun and Li, Ruiyang and Wang, Ruixin and Stanojcic, Nick and Li, Ji-Peng Olivia and Long, Erping and Wang, Jinghui and Zhang, Xiayin and Li, Jianbin and others},
  journal={International Journal of Surgery},
  volume={104},
  pages={106740},
  year={2022},
  publisher={Elsevier}
}

@inproceedings{pissas2021effective,
  title={Effective semantic segmentation in cataract surgery: What matters most?},
  author={Pissas, Theodoros and Ravasio, Claudio S and Da Cruz, Lyndon and Bergeles, Christos},
  booktitle={International Conference on Medical Image Computing and Computer-Assisted Intervention},
  pages={509--518},
  year={2021},
  organization={Springer}
}

@ARTICLE{cholec80,
  author={Twinanda, Andru P. and Shehata, Sherif and Mutter, Didier and Marescaux, Jacques and de Mathelin, Michel and Padoy, Nicolas},
  journal={IEEE Transactions on Medical Imaging}, 
  title={EndoNet: A Deep Architecture for Recognition Tasks on Laparoscopic Videos}, 
  year={2017},
  volume={36},
  number={1},
  pages={86-97},
  doi={10.1109/TMI.2016.2593957}}

@article{fang2022global,
  title={Global--local multi-stage temporal convolutional network for cataract surgery phase recognition},
  author={Fang, Lixin and Mou, Lei and Gu, Yuanyuan and Hu, Yan and Chen, Bang and Chen, Xu and Wang, Yang and Liu, Jiang and Zhao, Yitian},
  journal={BioMedical Engineering OnLine},
  volume={21},
  number={1},
  pages={82},
  year={2022},
  publisher={Springer}
}

@inproceedings{czempiel2020tecno,
  title={Tecno: Surgical phase recognition with multi-stage temporal convolutional networks},
  author={Czempiel, Tobias and Paschali, Magdalini and Keicher, Matthias and Simson, Walter and Feussner, Hubertus and Kim, Seong Tae and Navab, Nassir},
  booktitle={International conference on medical image computing and computer-assisted intervention},
  pages={343--352},
  year={2020},
  organization={Springer}
}

@article{cicinelli2023cataracts,
  title={Cataracts},
  author={Cicinelli, Maria Vittoria and Buchan, John C and Nicholson, Maneck and Varadaraj, Varshini and Khanna, Rohit C},
  journal={The Lancet},
  volume={401},
  number={10374},
  pages={377--389},
  year={2023},
  publisher={Elsevier}
}

@article{hira2022video,
  title={Video-based assessment of intraoperative surgical skill},
  author={Hira, Sanchit and Singh, Digvijay and Kim, Tae Soo and Gupta, Shobhit and Hager, Gregory and Sikder, Shameema and Vedula, S Swaroop},
  journal={International journal of computer assisted radiology and surgery},
  volume={17},
  number={10},
  pages={1801--1811},
  year={2022},
  publisher={Springer}
}

@article{kim2019objective,
  title={Objective assessment of intraoperative technical skill in capsulorhexis using videos of cataract surgery},
  author={Kim, Tae Soo and O’Brien, Molly and Zafar, Sidra and Hager, Gregory D and Sikder, Shameema and Vedula, S Swaroop},
  journal={International journal of computer assisted radiology and surgery},
  volume={14},
  number={6},
  pages={1097--1105},
  year={2019},
  publisher={Springer}
}

@inproceedings{zhao2022trasetr,
  title={Trasetr: track-to-segment transformer with contrastive query for instance-level instrument segmentation in robotic surgery},
  author={Zhao, Zixu and Jin, Yueming and Heng, Pheng--Ann},
  booktitle={2022 International conference on robotics and automation (ICRA)},
  pages={11186--11193},
  year={2022},
  organization={IEEE}
}

@article{giap2025catskill,
  title={CatSkill: Artificial Intelligence-Based Metrics for the Assessment of Surgical Skill Level from Intraoperative Cataract Surgery Video Recordings},
  author={Giap, Binh Duong and Ballouz, Dena and Srinivasan, Karthik and Lustre, Jefferson and Likosky, Keely and Mahmoud, Ossama and Mian, Shahzad I and Tannen, Bradford L and Nallasamy, Nambi},
  journal={Ophthalmology Science},
  volume={5},
  number={4},
  pages={100764},
  year={2025},
  publisher={Elsevier}
}

@article{ravi2024sam,
  title={Sam 2: Segment anything in images and videos},
  author={Ravi, Nikhila and Gabeur, Valentin and Hu, Yuan-Ting and Hu, Ronghang and Ryali, Chaitanya and Ma, Tengyu and Khedr, Haitham and R{\"a}dle, Roman and Rolland, Chloe and Gustafson, Laura and others},
  journal={arXiv preprint arXiv:2408.00714},
  year={2024}
}

@article{daneshgar2021visual,
  title={Visual intelligence: prediction of unintentional surgical-tool-induced bleeding during robotic and laparoscopic surgery},
  author={Daneshgar Rahbar, Mostafa and Ying, Hao and Pandya, Abhilash},
  journal={Robotics},
  volume={10},
  number={1},
  pages={37},
  year={2021},
  publisher={MDPI}
}
}

\clearpage

\setcounter{section}{0}
\maketitlesupplementary





\section{Final classification}
For each complication, we aggregate the segment-level predictions from the VLM to obtain a final video-level classification.
\begin{itemize}
    \item \textbf{Iris Prolapse}: Iris prolapse typically presents with clear and distinctive visual cues. Thus, if \textit{any} segment in a video is classified as positive, the entire video is labeled positive for iris prolapse.
    \item \textbf{Posterior Capsular Rupture (PCR):} PCR signatures can visually resemble other artifacts such as residual cortex, making segment-level classification inherently ambiguous. To minimize false negatives, we treat both \emph{positive} and \emph{unclear} predictions as indicative of PCR. Accordingly, a video is classified as positive if at least one segment receives either label.
    \item \textbf{Vitreous Loss}: We utilize a more conservative aggregation strategy to reduce false positives in vitreous loss classification. A video is classified as positive if at least one segment receives a \textit{high-confidence positive prediction}, or at least two segments receive \textit{medium-confidence positive} predictions.
\end{itemize}

\section{Prompts}
Below we provide the prompts used by the VLM to classify each of the three complications.
\begin{figure*}[t]
\centering

\begin{tcolorbox}[colback=gray!5,colframe=black!50,title=Iris Prolapse Detection Prompt]
\textbf{You are an expert ophthalmologist analyzing cataract surgery frames for iris prolapse detection.}

\textbf{IRIS PROLAPSE DEFINITION:}
\begin{itemize}
    \item Iris tissue protruding through the corneal incision
    \item Iris tissue extending beyond the normal eye boundary
    \item Dark iris tissue visible outside the pupil/iris border
    \item Tissue herniation through surgical wounds
\end{itemize}

\textbf{TASK:} Analyze this cataract surgery frame and determine:
\begin{enumerate}
    \item Is there iris prolapse present? (YES/NO)
    \item Confidence level (High/Medium/Low)
    \item Describe what you observe in the iris region
\end{enumerate}

Please be very specific about what you see in the iris area and whether any tissue appears to be protruding or displaced.

\textbf{RESPOND IN THIS FORMAT:}

\texttt{PROLAPSE\_DETECTED: [YES/NO]}\\
\texttt{CONFIDENCE: [High/Medium/Low]}\\
\texttt{DESCRIPTION: [Your detailed observation]}
\end{tcolorbox}

\end{figure*}

\begin{figure*}
\centering
\begin{tcolorbox}[colback=gray!5,colframe=black!50,title=PCR Detection Prompt]
\textbf{SYSTEM ROLE:} \\
You are an expert ophthalmologist and cataract surgery video reviewer. Classify a single frame (or short sequence) for signs of Posterior Capsule Rupture (PCR).

\vspace{4pt}
\textbf{PRIMARY GUIDING PRINCIPLE:}
\begin{itemize}
    \item Avoid false positives. Only classify ``Yes'' when the posterior capsule disruption is clearly visible or supported by strong secondary evidence.
    \item If capsule visibility is poor, ambiguous, or partially obscured (e.g., bubbles, glare, blur, instrument), prefer ``Unsure'' — interpreted as possible PCR with low confidence.
    \item Only use ``No'' when the capsule contour and posterior plane appear clearly continuous and stable.
\end{itemize}

\textbf{PRIMARY RULES:}
\begin{itemize}
    \item The posterior capsule normally appears as a smooth, taut, transparent plane behind the lens material.
    \item PCR requires visible or strongly implied loss of this contour — a distinct discontinuity, irregular tear, or missing segment.
    \item A bright posterior cavity or visible red reflex through the lens plane supports PCR only if the capsule edge is discontinuous or missing.
    \item Residual cortex folds, transient fluid waves, or smooth bulges are \textbf{not sufficient} for PCR unless the contour breaks or red reflex is sharply exposed.
    \item Instrument or fluid artifacts should not be mistaken for rupture unless the deformation persists or reveals a cavity.
\end{itemize}

\textbf{KEY DEFINITIONS:}
\begin{itemize}
    \item \textbf{Posterior capsule:} Transparent, faint circular membrane enclosing the lens.
    \item \textbf{PCR:} Sharp, irregular tear or absence of capsule continuity exposing deeper cavity or bright posterior reflex.
    \item \textbf{Residual cortex:} Smooth, curved, concentric opacities preserving capsule shape.
    \item \textbf{Lens fragment drop:} Nuclear/cortical fragment posterior to the capsule plane.
    \item \textbf{Red reflex abnormality:} Sharply localized posterior glow inconsistent with an intact capsule.
    \item \textbf{Deepened anterior chamber:} Posterior shift of iris–lens diaphragm suggesting capsular loss.
\end{itemize}

\textbf{SECONDARY SIGNS} (raise likelihood \textit{only} when capsule discontinuity is suspected):
\begin{itemize}
    \item Posterior cavity or glow consistent with loss of support.
    \item Posterior displacement of nuclear fragments.
    \item Deepened anterior chamber.
    \item Radial tear or flap-like capsule edges.
\end{itemize}

\end{tcolorbox}
\end{figure*}

\begin{figure*}[t]
\centering
\begin{tcolorbox}[colback=gray!5,colframe=black!50]

\textbf{OUTPUT FORMAT (valid JSON):}
\begin{verbatim}
{
  "label": "Yes" | "Unsure" | "No",
  "confidence": "High" | "Medium" | "Low",
  "reasons": [
    "Visual-only statements explaining decision"
  ],
  "observations": {
    "posterior_capsule_continuity": "intact" | "discontinuous" |
                                   "absent" | "unclear" | "unassessable",
    "radial_tears_or_folds": true | false,
    "lens_fragment_drop": true | false,
    "chamber_depth_change": true | false,
    "red_reflex_abnormality": true | false,
    "instrument_or_fluid_action": "none" | "present",
    "view_obstruction": ["glare", "bubbles", "blur", "partial_view"]
  },
  "notes": "Comment on ambiguity, lighting, or visibility."
}
\end{verbatim}

\textbf{CLASSIFICATION LOGIC (specificity-prioritized):}
\begin{itemize}
    \item Poor-quality / non-diagnostic frame → \texttt{"No"}.
    \item Capsule = intact and no secondary signs → \texttt{"No"} (High).
    \item Capsule = discontinuous/absent:
    \begin{itemize}
        \item With red reflex abnormality, fragment drop, or chamber depth change → \texttt{"Yes"} (High).
        \item Subtle/partial discontinuity → \texttt{"Yes"} (Medium).
    \end{itemize}
    \item Capsule = unclear:
    \begin{itemize}
        \item Mild red reflex abnormality / possible cavity → \texttt{"Unsure"} (Low).
        \item Folds without definite break → \texttt{"Unsure"} (Low).
        \item Otherwise → \texttt{"No"} (Medium).
    \end{itemize}
    \item Instrument/fluid distortion smooth or transient → \texttt{"No"} (Medium).
    \item ``Unsure'' is for ambiguous or borderline PCR cues.
\end{itemize}

\textbf{STYLE NOTES:}
\begin{itemize}
    \item Be concise, visual, and specific to the frame.
    \item Focus on contour geometry and lighting consistency.
    \item Avoid over-interpreting artifacts or reflections.
    \item When uncertain, explicitly state PCR is \textit{possible but not definitive}.
\end{itemize}

\end{tcolorbox}
\end{figure*}

\begin{figure*}
\centering
\begin{tcolorbox}[colback=gray!5,colframe=black!50,title=Vitreous Loss Detection Prompt]
\textbf{SYSTEM ROLE:}\\
You are an expert ophthalmologist and cataract surgery video reviewer. Classify a single frame for signs of Vitreous Loss (VL).

\vspace{4pt}
\textbf{PRIMARY GUIDING PRINCIPLE (sensitivity-first):}
\begin{itemize}
    \item Prefer not to miss VL. When evidence is ambiguous, prefer ``Unsure'' (or ``Yes'' with lower confidence) rather than a definitive ``No''.
\end{itemize}

\textbf{PRIMARY RULES:}
\begin{itemize}
    \item Normal pupil is round.
    \item If pupil shape is elongated (ellipse) or tear-drop (with or without sharp apex), and no clear instrument or OVD jet is visually causing the distortion $\rightarrow$ candidate for VL.
    \item Round pupil $\rightarrow$ ``No'' unless strong secondary signs indicate VL.
    \item Elongated pupil \textit{with} iris prolapse $\rightarrow$ ``Unsure'' (document prolapse).
    \item If the pupil is obstructed (bubbles or instrument) $\rightarrow$ ``Unsure''.
\end{itemize}

\textbf{KEY DEFINITIONS:}
\begin{itemize}
    \item \textbf{Pupil peaking/elongation:} Any deviation from circular toward ellipse or tear-drop; note apex sharpness and margin continuity.
    \item \textbf{Iris prolapse:} Iris tissue protruding through incision or beyond normal boundary.
    \item \textbf{PCR indicators:} Bright red/hemorrhagic glow behind pupil OR visible posterior capsule rent.
    \item \textbf{Instrument-induced distortion:} Any tool, OVD jet, irrigation/aspiration, or IOL insertion contacting iris/pupil. Not always exclusion of VL.
    \item \textbf{OVD jet vs vitreous:} OVD is smooth/continuous/transient; vitreous is thin, jelly-like, fibrillar, persistent, and refracts light.
\end{itemize}

\textbf{SECONDARY SIGNS (strengthen ``Yes''):}
\begin{itemize}
    \item Bright red/hemorrhagic glow in pupil background.
    \item Jelly-like fibrillar strands attached to wound, across pupil, or bridging instruments.
    \item Microbubbles along clear strands or refracting points.
    \item Instruments interacting with stringy, gelatinous material.
\end{itemize}

\textbf{OUTPUT FORMAT (valid JSON):}

\begin{verbatim}
{
  "label": "Yes" | "Unsure" | "No",
  "confidence": "High" | "Medium" | "Low",
  "reasons": [
      "Visual-only statements explaining decision"
  ],
  "observations": {
    "pupil_shape": "round" | "ellipse" | "tear-drop",
    "pupil_apex_sharp": true | false,
    "instrument_or_fluid_action": "none" | "instrument" | "OVD" | "IA" | "IOL",
    "ovd_jet_visible": true | false,
    "iris_prolapse_signs": true | false,
    "pcr_signs": {
      "pc_rent_visible": true | false,
      "retro_red_glow": true | false
    },
    "vitreous_strands_visible": true | false,
    "wound_attachment_of_strands": true | false,
    "frame_quality_issues": ["glare","blur","partial_view"]
  }
}
\end{verbatim}

\end{tcolorbox}
\end{figure*}

\begin{figure*}[t]
\centering
\begin{tcolorbox}[colback=gray!5,colframe=black!50]

\textbf{CLASSIFICATION LOGIC (sensitivity-first):}
\begin{itemize}
    \item Severe frame-quality issues reduce confidence by one level.
    \item \textbf{If pupil is round:}
    \begin{itemize}
        \item If any secondary signs (strands, PCR glow) $\rightarrow$ ``Yes''.
        \item Else $\rightarrow$ ``No''.
    \end{itemize}

    \item \textbf{If pupil is ellipse or tear-drop:}
    \begin{itemize}
        \item \textbf{If no instrument/fluid action:}
        \begin{itemize}
            \item If prolapse visible $\rightarrow$ ``Unsure''.
            \item Else $\rightarrow$ ``Yes'' (High if strands/PCR signs; Medium otherwise).
        \end{itemize}

        \item \textbf{If instrument/fluid action present:}
        \begin{itemize}
            \item If OVD jet visible and distortion looks smooth/transient (no strands) $\rightarrow$ ``No'' (Medium).
            \item If strands, PCR glow, or wound-attached material visible $\rightarrow$ ``Yes'' (Medium--High).
            \item If neither VL nor OVD pattern is clearly visible $\rightarrow$ ``Unsure'' (Low--Medium). 
        \end{itemize}
    \end{itemize}

    \item Iris prolapse \textit{with} strands or PCR signs $\rightarrow$ ``Yes''.
    \item Always provide specific visual cues (e.g., “thin fibrillar strands 7--10 o'clock”, “retro-red glow centered”).
\end{itemize}

\textbf{STYLE:}
\begin{itemize}
    \item Be concise, visual, and strictly tied to what is seen.
    \item Avoid speculation; prioritize sensitivity (avoid false negatives).
    \item When ambiguous, recommend temporal review.
\end{itemize}

\end{tcolorbox}
\end{figure*}

\end{document}